\documentclass[letterpaper]{article}
\usepackage[preprint]{aaai2027}
\usepackage[hyphens]{url}
\usepackage{graphicx}
\usepackage{natbib}
\usepackage{caption}
\usepackage{amsmath}
\usepackage{amssymb}
\usepackage{amsthm}
\newtheoremstyle{lure}{2pt}{2pt}{\itshape}{0pt}{\bfseries}{.}{.5em}{}
\theoremstyle{lure}
\newtheorem{theorem}{Theorem}
\newtheorem{proposition}[theorem]{Proposition}
\usepackage{dsfont}

\usepackage{booktabs}
\usepackage{colortbl}
\definecolor{bestbg}{HTML}{D6E4F4}
\definecolor{secbg}{HTML}{EEF3FA}
\newcommand{\tb}[1]{\cellcolor{bestbg}$\mathbf{#1}$}
\newcommand{\ts}[1]{\cellcolor{secbg}$\underline{#1}$}

\usepackage{algorithm}
\usepackage{algorithmic}

\title{The Chase Is the Curriculum, the Capture Anchors the Credit:\\ Pursuit-Evasion Self-Play for Zero-Data LLM Reasoning}

\author{
    Jing Yu\textsuperscript{\rm 1},
    Shengchao Chen\textsuperscript{\rm 2}\corresponding,
    Yiyun Tan\textsuperscript{\rm 1}
}
\affiliations{
    \textsuperscript{\rm 1}College of Computer and Information Engineering,
    Xi'an University of Architecture and Technology\\
    \textsuperscript{\rm 2}Australian Artificial Intelligence Institute,
    University of Technology Sydney\\
    yujing1203@xauat.edu.cn, shengchao.chen.uts@gmail.com, yytanmax@gmail.com
}

\begin{document}
\maketitle

\begin{abstract}
Reinforcement learning with verifiable rewards has become the dominant recipe for improving large language model reasoning, yet it presumes large human-curated task collections. Zero-data self-play removes this dependency, but existing methods vet learnability only by probing candidates and rejecting post hoc, never learning where along an environment's difficulty axis to place a task, and credit the solver with sparse terminal rewards alone. We recast zero-data self-play as a \emph{pursuit-evasion game}: in \textbf{LURE}, an LLM \emph{evader} positions tasks along each environment's difficulty axis to stay one step ahead of a planner-executor \emph{pursuer} that hunts it down through verifiable interaction. The evader is trained on a \emph{capture-frontier} reward that peaks when the solver captures it on exactly half of its rollouts, turning barely catchable into a learned positioning strategy rather than a hand-tuned rejection band. The pursuer earns \emph{capture-anchored dense process credit}, in which monotone verifier progress is group-normalized jointly with the terminal capture under a round-anchored KL that keeps the co-evolution stable. Across three verifiable reasoning environments and three backbone families, \textbf{LURE} outperforms advanced baselines under unified/specialist settings, while the unified model attains stronger aggregate OOD zero-shot accuracy than all trained baselines across nine held-out benchmarks from three task families.

\end{abstract}

\section{Introduction}

Large language models are increasingly deployed on multi-step verifiable reasoning tasks, including multi-hop question answering~\citep{phantomwiki2024}, instruction following~\citep{zhou2023ifeval}, and logic-grid deduction~\citep{lin2025zebralogic}, where success requires planning and executing several intermediate steps. The prevailing route to such abilities is reinforcement learning with verifiable rewards (RLVR), which optimizes the policy directly against programmatic checkers rather than a learned reward model \citep{deepseekmath,deepseekr1,tulu3}. Its effectiveness, however, is bounded by the supply of human-curated tasks with gold verification, which is expensive to build and quickly saturated by strong models \citep{serl2025}. Self-evolving training promises to lift this ceiling: the model generates its own curriculum and improves from verifiable feedback with \emph{zero} external data \citep{rzero2025,absolutezero2025}.

\begin{figure*}[tbh]
  \centering
  \includegraphics[width=.82\textwidth]{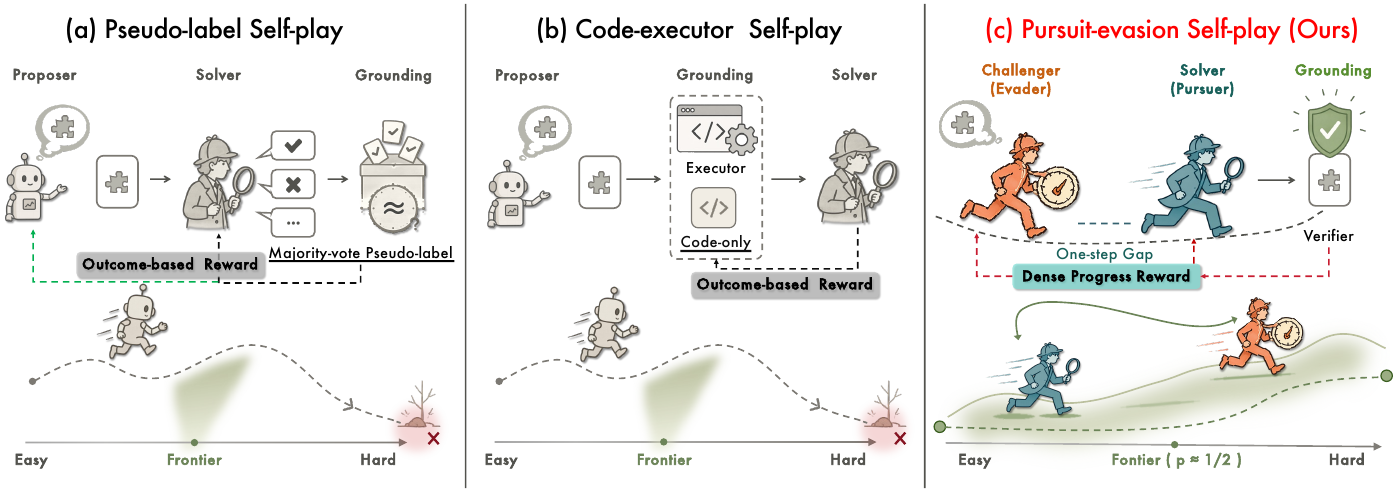}
  \caption{Zero-data self-play paradigms. \emph{(a)} Pseudo-label self-play uses majority-vote labels~\citep{rzero2025}, and \emph{(b)} code-executor self-play uses program execution~\citep{absolutezero2025}. Both author tasks free-form, vetted post hoc under outcome-only supervision. \emph{(c)} \textbf{LURE} adapts the challenger toward solver's capture frontier and gives the pursuer dense verifier-based credit.}
  \vspace{-4pt}
  \label{fig:hero}
\end{figure*}

Existing zero-data self-play, whether grounded in majority-vote pseudo-labels~\citep{rzero2025} or a code executor~\citep{absolutezero2025} (\textbf{Fig.~\ref{fig:hero}}a,b), gives its curriculum no explicit difficulty axis: a proposer authors candidate tasks in text space, and learnability is enforced only \emph{post hoc}, by probing each candidate with solver rollouts and rejecting those whose empirical solve rate falls outside a fixed band \citep{rzero2025}. Task placement therefore stays implicit in the authored text, and the hand-set band edges never adapt as the solver's competence changes. When that proposal is mismatched with the solver, rejection sampling degenerates into paying rollout cost only to discard almost everything.

The same non-adaptive curriculum leaves the solver under-informed. A multi-step episode returns a single terminal reward, so the trajectory is credited as one block, and under group-relative policy optimization the gradient vanishes whenever every rollout of a task fails \citep{zerorewardbarrier2025}, precisely the case for hard, self-generated tasks. Densifying this signal with a process reward model reintroduces the external supervision that self-play is meant to eliminate \citep{lightman2024lets,wang2024mathshepherd}, and the round-by-round drift of the task distribution makes optimization non-stationary and prone to collapse. What is missing is a loop that learns from the verifier alone: a challenger trained to place tasks at the solver's moving frontier, and dense credit derived, without annotation, from the environment's own progress signal, stabilized across co-evolution.

To address these challenges, this paper formulates zero-data self-play as a \emph{pursuit-evasion game} (\textbf{Fig.~\ref{fig:hero}}c), reminiscent of Scotland Yard and of cops-and-robbers games on graphs~\citep{bonato2011cops}, in which an LLM challenger (the evader) learns where to \emph{position} tasks and a planner-executor solver (the pursuer) learns to hunt them down, co-evolving on verifiable natural-language reasoning environments. The evader is optimized with GRPO on a \emph{capture-frontier} reward that peaks when the solver captures a task on half of its rollouts, minus a difficulty-signature repetition penalty, so ``stay exactly one step ahead of the pursuer'' becomes a positioning strategy the challenger \emph{learns}, rather than a hand-tuned rejection band applied after blind sampling. In turn, the pursuer is optimized with \emph{capture-anchored dense process credit}, in which each executor step earns the monotone verifier-progress it adds, with zero-progress steps paying a small redundancy cost, jointly group-normalized with the terminal capture, under a round-anchored KL applied to both players so that neither escapes the other's reach within a round. Instantiated over three heterogeneous environments with per-role weights shared across all of them, this chase turns verifier feedback into both players' training signal without any human data. We call the method \textbf{LURE}: the evader acts as a learned lure, holding each task barely catchable so that it pulls the solver up to its own frontier rather than fleeing beyond reach. Our contributions are fourfold:
\begin{itemize}
\item We formulate zero-data self-play as a pursuit-evasion game and, to our knowledge, are the first to \emph{co-train} an LLM difficulty-selection challenger with a multi-turn planner-executor solver on verifiable NL reasoning.
\item We propose a capture-frontier challenger objective that turns the solver's group capture statistics into a training signal for difficulty positioning, replacing post-hoc rejection with a learned curriculum policy.
\item We introduce a capture-anchored dense process credit from the environment verifier alone, with round-anchored KL stabilization applied to both players.
\item Across three verifiable reasoning environments and backbone families, \textbf{LURE} outperforms advanced baselines under unified/specialist settings, attaining stronger aggregate OOD zero-shot accuracy across nine benchmarks.
\end{itemize}

\section{Related Work}

\paragraph{Verifiable Multi-Step Reasoning.}
Multi-step reasoning has advanced by making intermediate computation explicit and verifiable. Chain-of-thought exposes reasoning paths~\citep{wei2022chain}, while step-level scoring improves both training and inference beyond final-answer supervision~\citep{lightman2024lets}. Reinforcement learning with verifiable rewards turns programmatic checkers into optimization targets, with group-relative policy optimization now standard for reasoning LLMs~\citep{deepseekmath,deepseekr1,sheng2024hybridflow}. For multi-step tasks, agents interleave decomposition, grounded action, and self-correction~\citep{yao2023react,shinn2023reflexion}, while process reward models densify terminal rewards but typically require step annotations that zero-data methods seek to avoid~\citep{wang2024mathshepherd}. These advances also motivate programmatically verifiable benchmarks for instruction following~\citep{zhou2023ifeval}, logic-grid deduction~\citep{lin2025zebralogic}, and multi-hop relational question answering~\citep{phantomwiki2024}. This line of work improves \emph{solving} but leaves task generation open.

\paragraph{LLM Self-Evolving.}
Self-evolving training lets models generate their own curricula by bootstrapping supervision through synthesized instructions~\citep{wang2023selfinstruct} or outcome-filtered rationales~\citep{zelikman2022star}, or by framing training as a game between a proposer and a solver~\citep{spiral2025,absolutezero2025}. Most closely related, R-Zero co-evolves an LLM challenger that authors problem text, keeping candidates only when solve rates fall within a fixed hand-set band~\citep{rzero2025}, which costs a probe of solver rollouts per candidate. Separately, outcome-only group-relative training provides no learning signal when an entire group fails, known as the zero-reward barrier~\citep{zerorewardbarrier2025}. Beyond language models, adaptive task generation at the learner's frontier appears in asymmetric self-play~\citep{asymmetricselfplay}, goal generators targeting intermediate success rates~\citep{goalgan}, and regret-based environment design~\citep{paired}. Our setting instead considers natural-language tasks generated by an LLM proposer and supervised solely by a programmatic verifier. Building on the pursuit-evasion framing~\citep{bonato2011cops}, \textbf{LURE} trains the evader with a capture-frontier reward and the pursuer with dense verifier progress, using the same programmatic signal for both roles.

\section{Methodology}

\begin{figure*}[tbh]
  \centering
  \includegraphics[width=.78\textwidth]{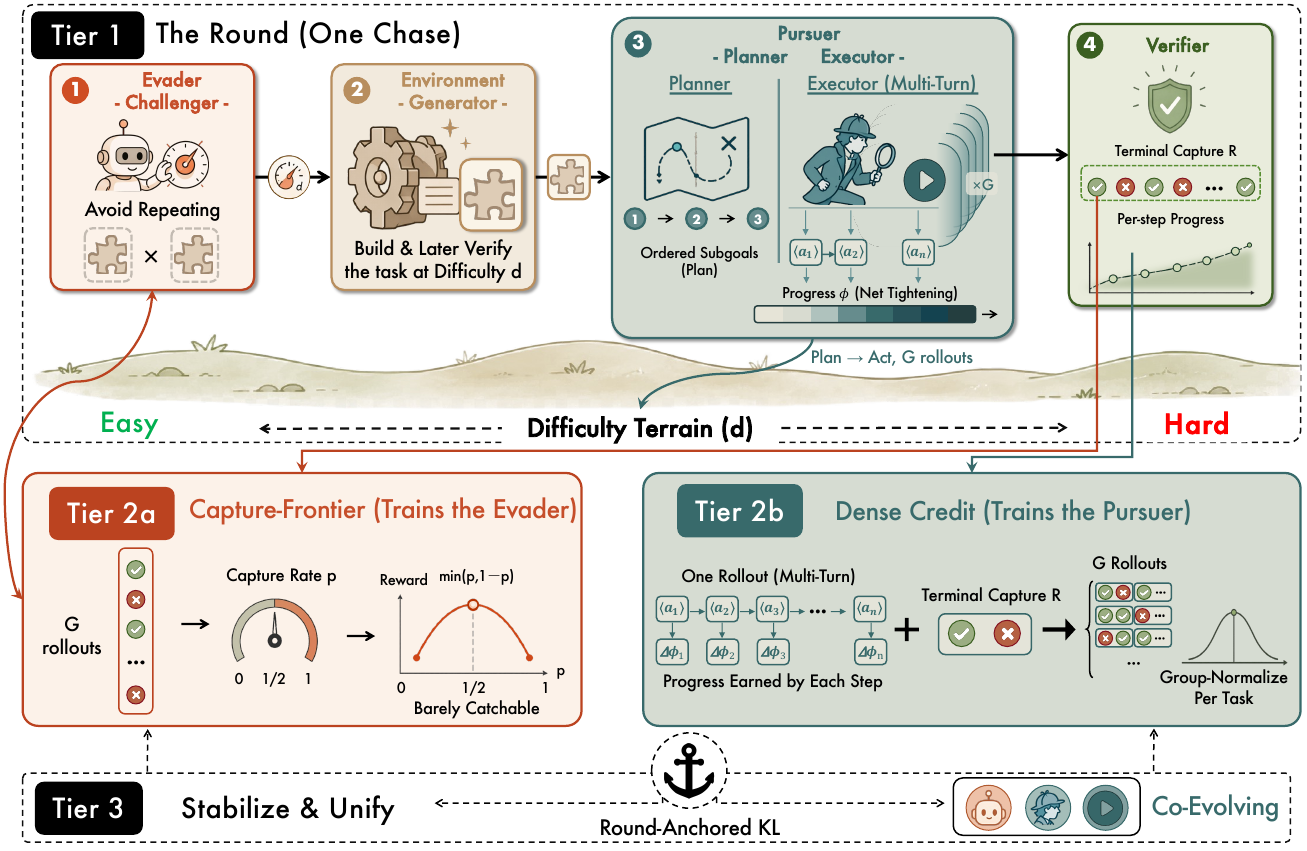}
  \caption{\textbf{Overview of LURE.}
\emph{Tier 1:} the evader selects a difficulty $d$ with a penalty for repeated signatures, the generator instantiates the task, the planner-executor pursuer runs $G$ multi-turn rollouts, and the verifier returns capture $R$ and stepwise progress $\phi$.
\emph{Tier 2a:} the evader optimizes $\min(p,1{-}p)$, peaking at the capture frontier $p=1/2$.
\emph{Tier 2b:} the pursuer receives task-wise group-normalized credit from $R$ and $\Delta\phi$.
\emph{Tier 3:} a round-anchored KL stabilizes all three policies.}
  \label{fig:overview}
  \vspace{-4pt}
\end{figure*}
\paragraph{Problem Formulation.}
We consider a family of verifiable language reasoning environments $\mathcal{E}$. Each environment $e$ provides a programmatic task generator $g_e(d,\omega)$ indexed by a difficulty parameter $d \in [0,1]$, a multi-turn interaction protocol, and a verifier that emits two signals per trajectory $\tau$: a binary terminal \emph{capture} $R(\tau) \in \{0,1\}$, which fires only when the task is solved in full, and a running progress $\phi_t \in [0,1]$, the best verified progress so far, with $\phi = 1$ at capture. We write $\Delta\phi_t$ for the nonnegative increment in best verified progress. We cast self-play as a two-player pursuit-evasion interaction: the \emph{evader} $\pi_C$ selects where on each environment's difficulty axis to place the next tasks, which the generator instantiates, and the \emph{pursuer} $\pi_S = (\pi_{\text{plan}}, \pi_{\text{exec}})$ is a planner-executor pair that hunts each task down, with empirical capture rate on task $x$ over $G$ rollouts
\begin{equation}
p(x) \;=\; \frac{1}{G}\sum_{g=1}^{G} \mathds{1}\!\left[R(\tau_g)=1\right].
\end{equation}
The two players optimize a coupled pair of objectives, stated as a pair rather than a minimax and carrying no equilibrium or convergence claim:
\begin{align}
\max_{\pi_C}\;\; & \mathbb{E}_{x}\!\left[\min\bigl(p(x),\,1-p(x)\bigr) - \rho(x)\right], \nonumber\\
\max_{\pi_S}\;\; & \mathbb{E}_{x \sim \pi_C}\,\mathbb{E}_{\tau}\!\left[R(\tau) + \lambda \sum_t \bigl(\Delta\phi_t - c_{red}\,\mathds{1}[\Delta\phi_t{=}0]\bigr)\right].
\end{align}
Both objectives are functions of the verifier alone, with no human-curated tasks, labels, or reward models.

\paragraph{Overview.}
\textbf{Fig.~\ref{fig:overview}} shows the workflow of \textbf{LURE}. A single round of the chase couples three co-evolving policies, the evader, the planner, and the executor, instantiated as three separate models. Each round opens by freezing a snapshot of all three, and those snapshots generate every sample the round collects. First, the evader emits $K$ candidate tasks per environment, each placed at a target difficulty, and the well-formed ones become the round's curriculum. Second, the pursuer attacks each task with $G$ plan-execute rollouts while the verifier records, per trajectory, the terminal capture $R$ and the per-step progress $\{\phi_t\}$. Third, capture statistics reposition the evader near the pursuer's frontier, capture and progress increments credit the executor, and the progress its plans realize credits the planner. Finally, each policy takes a single GRPO update on this round's data. Because every trajectory was generated by the round-start snapshots, the round is simultaneous-move, and each player faces the others' updates only in the next round.

\paragraph{Frontier-Seeking Evader.}
The evader learns \emph{where to run} rather than what the tasks contain, repositioning along the difficulty axis against the pursuer's measured competence. Each emitted task $x$ receives the capture-frontier reward
\begin{equation}
\label{eq:evader_reward}
r_C(x) =
\begin{cases}
\min\bigl(p(x),\, 1-p(x)\bigr) - \rho(x), & x \in \mathcal{V},\\[2pt]
-1 - \rho(x), & x \notin \mathcal{V},
\end{cases}
\end{equation}
where $\mathcal{V}$ is the set of well-formed emissions the pursuer attacked, and the repetition penalty $\rho(x) = (n_{sig}(x) - 1)/N$ grows with the number of the round's $N$ emissions that share $x$'s signature, so the evader cannot collapse onto one signature. Signatures are coarse and per-environment: hop count and population for multi-hop question answering, constraint count for instruction following, grid dimensions for logic-grid deduction. The reward peaks at $p = \tfrac{1}{2}$, so the evader's optimum is to remain \emph{barely catchable}, neither escaping ($p{=}0$) nor trivially caught ($p{=}1$). This is also where the capture signal is most informative, since binary capture gives the terminal reward a within-group variance of $p(1-p)$, and a task off the frontier returns near-identical outcomes across its group and a vanishing group-normalized advantage. Rejection filtering, by contrast, discards finished candidates but never steers where the next ones are placed. Evader advantages are group-normalized over the samples sharing a prompt slot. The generation prompt also carries an initial difficulty hint and a per-difficulty summary of pursuer competence, so this scaffolding shapes the frontier behavior alongside the reward.

\paragraph{Dense-Credit Pursuer.}
Terminal capture is sparse: on hard self-generated tasks, whole rollout groups fail and a purely terminal group-relative signal vanishes. The pursuer's dense credit therefore comes from the verifier's own step progress. For each executor turn $t$ of trajectory $\tau$ we form a single raw scalar
\begin{equation}
\label{eq:rt}
r_t \;=\; R(\tau)\,\mathds{1}\!\left[t = T_\tau\right] \;+\; \lambda\left(\Delta\phi_t - c_{red}\,\mathds{1}\!\left[\Delta\phi_t = 0\right]\right),
\end{equation}
and standardize these scalars jointly over the group $\mathcal{G}(x)$ of all executor turns across the $G$ rollouts of $x$:
\begin{equation}
A_t \;=\; z_{\mathcal{G}(x)}\!\left(r_t\right).
\end{equation}
Three design choices matter. First, the terminal reward is credited once, on the last executor turn, since replicating it across turns would bias the group statistics toward long trajectories. Second, the zero-progress cost is charged only when $\Delta\phi_t = 0$: an unconditional constant would be exactly cancelled by the shift invariance of group-normalization, so the indicator separates stalling from progressing whenever a group contains both. Third, because $\phi_t$ is best-so-far, $\Delta\phi_t \ge 0$ and $\sum_t \Delta\phi_t$ telescopes to the trajectory's peak verified progress, so a regression receives zero dense credit rather than a penalty. When an entire group fails, the terminal term is identically zero and the dense term still carries gradient wherever step progress varies within the group. The planner receives its own credit: each rollout is scored by the stage potentials $\Phi_i$ of the executor segments it governed,
\begin{equation}
\label{eq:planner_credit}
r^{\mathrm{plan}}(\tau) = \sum_{i} \Phi_i,
\end{equation}
group-normalized over the same task group with no terminal term, so a plan is judged by how far its execution advanced rather than by whether capture occurred.

\paragraph{Stabilized Co-Evolution.}
The chase is non-stationary by construction. The task distribution the pursuer faces, and the pursuer the evader probes, both shift from round to round, so an update regularized toward a stale reference is pulled away from the distribution that produced its data. We therefore formulate the round as a proximal co-evolution step: the evader, planner, and executor each take a single GRPO update under a KL penalty to their own round-start snapshot. That snapshot acts as behavior policy, KL reference, and proximal center, aligning the regularizer with the distribution that generated the samples and limiting within-round drift. Without it we observed format and mode collapse. Re-anchoring at the next round preserves that control while letting the joint policy trajectory adapt across rounds.

\paragraph{Optimization Objective.}
All three roles are trained with the same clipped GRPO objective~\citep{deepseekmath,ppo} and round-anchored KL, differing only in their advantages. Let $\bar\pi_r$ denote the round-start snapshot of role $r$ and $\varrho_r = \pi_{\theta_r}(y \mid s)/\bar\pi_r(y \mid s)$ the token-level ratio to it. Each role maximizes the objective below:
\begin{equation}
\begin{aligned}
\mathcal{J}_r(\theta_r) = \; & \mathbb{E}\!\left[\min\bigl(\varrho_r A^r,\, \mathrm{clip}(\varrho_r, 1{-}\epsilon, 1{+}\epsilon)\, A^r\bigr)\right] \\
& - \beta\, \mathrm{KL}\bigl(\pi_{\theta_r} \,\|\, \bar\pi_r\bigr),
\end{aligned}
\end{equation}
for $r \in \{C, \mathrm{plan}, \mathrm{exec}\}$. Each role's advantage is its group-normalized reward: $A^C = z_{\text{slot}}(r_C)$ for the evader, $A^{\mathrm{exec}}_t = z_{\mathcal{G}(x)}(r_t)$ for the executor, and $A^{\mathrm{plan}}_\tau = z_{\mathcal{G}(x)}\bigl(r^{\mathrm{plan}}(\tau)\bigr)$ for the planner, with every token carrying the advantage of its turn or rollout. \textbf{Alg.~\ref{alg:round}} shows the workflow.

\begin{algorithm}[t]
\caption{One round of \textbf{LURE}. \textbf{Alg.~2} in \textbf{App.~A} writes out every step.}
\label{alg:round}
\begin{algorithmic}[1]
\REQUIRE policies $\pi_C,\pi_{\text{plan}},\pi_{\text{exec}}$, environments $\mathcal{E}$, candidates $K$, rollouts $G$
\ENSURE updated policies $\pi_C,\pi_{\text{plan}},\pi_{\text{exec}}$
\STATE round-start snapshots $\bar\pi_C,\bar\pi_{\text{plan}},\bar\pi_{\text{exec}} \gets \pi_C,\pi_{\text{plan}},\pi_{\text{exec}}$
\STATE buffers $\mathcal{D}_C,\mathcal{D}_{\text{plan}},\mathcal{D}_{\text{exec}} \gets \emptyset$, task pool $\mathcal{X} \gets \emptyset$
\FOR{each environment $e \in \mathcal{E}$}
    \STATE $\mathcal{X} \gets \mathcal{X} \cup \{\text{valid tasks from } K \text{ of } \bar\pi_C \text{ through } g_e\}$
\ENDFOR
\FOR{each task $x \in \mathcal{X}$}
    \STATE roll out $G$ trajectories with $(\bar\pi_{\text{plan}},\bar\pi_{\text{exec}})$, scored by the verifier
    \STATE $p(x) \gets \frac{1}{G}\sum_{g=1}^{G}\mathds{1}[R(\tau_g){=}1]$
    \STATE add the evader sample for $x$ to $\mathcal{D}_C$ (Eq.~\ref{eq:evader_reward})
    \STATE add the executor turns to $\mathcal{D}_{\text{exec}}$ (Eq.~\ref{eq:rt})
    \STATE add the planner turns to $\mathcal{D}_{\text{plan}}$ (Eq.~\ref{eq:planner_credit})
\ENDFOR
\STATE update $\pi_C$ by GRPO on $\mathcal{D}_C$ with KL anchor $\bar\pi_C$
\STATE update $\pi_{\text{plan}}$ by GRPO on $\mathcal{D}_{\text{plan}}$ with KL anchor $\bar\pi_{\text{plan}}$
\STATE update $\pi_{\text{exec}}$ by GRPO on $\mathcal{D}_{\text{exec}}$ with KL anchor $\bar\pi_{\text{exec}}$
\STATE \textbf{return} $\pi_C,\pi_{\text{plan}},\pi_{\text{exec}}$
\end{algorithmic}
\end{algorithm}

\begin{table*}[tbh]
  \centering
  \resizebox{.96\textwidth}{!}{%
  \begin{tabular}{l cccc | cccc}
    \toprule
    & \multicolumn{4}{c}{\textit{Unified model}} & \multicolumn{4}{c}{\textit{Per-environment specialists}}\\
    \cmidrule(lr){2-5}\cmidrule(lr){6-9}
    & \textbf{PhantomWiki} & \textbf{IFEval} & \multicolumn{2}{c}{\textbf{ZebraLogic}} & \textbf{PhantomWiki} & \textbf{IFEval} & \multicolumn{2}{c}{\textbf{ZebraLogic}}\\
    \cmidrule(lr){4-5}\cmidrule(lr){8-9}
    Method & SUCCESS$\uparrow$ & Prompt Acc.$\uparrow$ & Puzzle Acc.$\uparrow$ & Cell Acc.$\uparrow$ & SUCCESS$\uparrow$ & Prompt Acc.$\uparrow$ & Puzzle Acc.$\uparrow$ & Cell Acc.$\uparrow$\\
    \midrule
    \multicolumn{9}{l}{\textbf{Qwen2.5-7B-Instruct}~\citep{qwen25}}\\
    Base                 & $25.0$ & \tb{63.0} & $11.6$ & $42.3$ & $25.0$ & \tb{63.0} & $11.6$ & $42.3$\\
    GRPO \citep{deepseekmath} & $43.3$ & $60.6$ & \ts{15.4} & \ts{45.8} & \ts{60.8} & \ts{62.8} & \ts{16.1} & \ts{46.5}\\
    GRPO-Zero            & $37.5$ & $58.4$ & $13.9$ & $44.2$ & $0.0$ & $54.9$ & $12.7$ & $43.1$\\
    R-Zero \citep{rzero2025} & \ts{50.8} & $57.9$ & $14.2$ & $43.6$ & $56.7$ & $60.1$ & $15.0$ & $44.8$\\
    \textbf{LURE (Ours)} & \tb{65.8} & \ts{61.2} & \tb{16.9} & \tb{47.2} & \tb{80.0} & \tb{63.0} & \tb{18.3} & \tb{48.6}\\
    \midrule
    \multicolumn{9}{l}{\textbf{Llama-3.1-8B-Instruct}~\citep{llama3}}\\
    Base                 & $0.0$ & $37.5$ & $12.8$ & $13.7$ & $0.0$ & $37.5$ & $12.8$ & $13.7$\\
    GRPO \citep{deepseekmath} & $0.8$ & $38.8$ & $14.1$ & \ts{27.9} & $2.5$ & \ts{43.4} & \ts{13.6} & $27.2$\\
    GRPO-Zero            & $0.8$ & $39.4$ & $13.0$ & $26.4$ & $0.8$ & $39.4$ & $12.9$ & $26.2$\\
    R-Zero \citep{rzero2025} & \ts{3.3} & \ts{42.5} & \ts{14.6} & $26.8$ & \ts{4.2} & $43.1$ & \ts{13.6} & \ts{28.3}\\
    \textbf{LURE (Ours)} & \tb{4.2} & \tb{44.5} & \tb{15.9} & \tb{29.5} & \tb{5.0} & \tb{44.7} & \tb{15.2} & \tb{29.9}\\
    \midrule
    \multicolumn{9}{l}{\textbf{Gemma-2-9B-it}~\citep{gemma2}}\\
    Base                 & $11.7$ & $60.4$ & $13.1$ & $38.4$ & $11.7$ & $60.4$ & $13.1$ & $38.4$\\
    GRPO \citep{deepseekmath} & $27.5$ & \ts{64.0} & \ts{15.7} & $40.2$ & \ts{40.0} & $62.5$ & \ts{15.7} & $40.3$\\
    GRPO-Zero            & $15.0$ & $61.2$ & $14.3$ & $39.9$ & $20.8$ & $60.4$ & $14.3$ & $39.6$\\
    R-Zero \citep{rzero2025} & \ts{34.2} & $62.5$ & $15.1$ & \ts{41.6} & $30.0$ & \ts{63.6} & $15.0$ & \ts{40.9}\\
    \textbf{LURE (Ours)} & \tb{40.0} & \tb{66.2} & \tb{18.2} & \tb{43.5} & \tb{42.5} & \tb{66.2} & \tb{17.6} & \tb{43.1}\\
    \bottomrule
  \end{tabular}}
  \caption{Main results (\%) on PhantomWiki, IFEval, and ZebraLogic across two settings. \colorbox{bestbg}{\textbf{Best}} and \colorbox{secbg}{\underline{second best}} per backbone.}
  \label{tab:main}
\end{table*}

\paragraph{Theoretical Analysis.}
Three exact properties of the design, proved in \textbf{App.~D}: the evader's optimum sits precisely where the pursuer's learning signal peaks, the repetition penalty makes curriculum collapse impossible, and the dense credit can never outrank capture.

\begin{theorem}[Frontier signal]
\label{thm:frontier}
Let $p$ be a task's capture probability and $A_1,\ldots,A_G$ the group-normalized advantages of its $G \ge 2$ binary capture outcomes. Then $\sum_g A_g^2 = (G-1)\mathds{1}[s(R)>0]$ pointwise, with $s(R)$ the sample standard deviation of the outcomes, so $\mathbb{E}[\sum_g A_g^2] = (G-1)(1 - p^G - (1-p)^G)$, uniquely maximized at $p = 1/2$, which is also the unique maximizer of $r_C$.
\end{theorem}

\begin{proposition}[No signature collapse]
\label{prop:collapse}
In the continuous relaxation the maximizer of Eq.~\ref{eq:evader_reward} over signature distributions is unique, with $m^\star_s = 1/k + (f_s - \bar f)/2$ on its support $S^\star$ of size $k$. Hence $k \ge 2$ and $|m^\star_s - m^\star_{s'}| \le 1/4$ on $S^\star$.
\end{proposition}

\begin{proposition}[Capture dominance]
\label{prop:dominance}
Best-so-far progress gives $\sum_t \Delta\phi_t = \phi_{T_\tau}$, so Eq.~\ref{eq:rt} totals $R(\tau) + \lambda\phi_{T_\tau} - \lambda c_{red}Z(\tau)$ over a trajectory, with $Z$ its stalled-turn count. If $\lambda c_{red}(T_{\max}-1) < 1$, every capturing trajectory strictly outranks every non-capturing one.
\end{proposition}

\section{Experiments}

\paragraph{Implementation Details.}
We used Qwen2.5-7B-Instruct, Llama-3.1-8B-Instruct, or Gemma-2-9B-it as our backbone models. All models were trained with verl-GRPO~\citep{sheng2024hybridflow,deepseekmath} under FSDP-2 on 16 $\times$ RTX~6000 Pro Blackwell $96$\,GB GPUs for $R=8$ rounds, using no external task data. Each round emits $48$ candidate tasks per environment ($8$ prompt slots $\times\,6$ samples) and attacks every surviving task with a group of $G=8$ pursuer rollouts. \textbf{LURE} sets the dense-credit weight $\lambda=0.25$ and zero-progress cost $c_{red}=0.05$, and runs the stabilized recipe on all three policies: round-anchored KL with $\beta=0.1$, learning rate $5\times10^{-6}$, and a single PPO epoch. Baselines use the optimizer settings from their own papers. The held-out protocol, covering prompts, verifiers, and decoding, is identical across methods, with full configurations, role prompts, and baseline constructions in \textbf{App.~A}.

\paragraph{Benchmarks \& Metrics.}
To evaluate multi-step verifiable reasoning, the setting the curriculum targets, we use three environments that carry their own programmatic verifier and need no human labels: \textbf{PhantomWiki}~\citep{phantomwiki2024} for relational inference, \textbf{IFEval}~\citep{zhou2023ifeval} for instruction following, and \textbf{ZebraLogic}~\citep{lin2025zebralogic} for logical deduction. Each doubles as a benchmark and is measured on a held-out split. PhantomWiki is multi-hop kinship question answering, scored as chain-verified exact match on $n=120$, where every hop must resolve and the first error ends the chain. IFEval is scored as per-prompt accuracy on its $541$ prompts under strict evaluation. ZebraLogic is scored on the official set of $1000$ puzzles, as puzzle accuracy, which demands every cell of the grid, and as cell accuracy, with every metric formalized in \textbf{App.~B}.

\paragraph{Main Results.}
We compare \textbf{LURE} with \emph{Base} (zero-shot), \emph{GRPO} trained on curated tasks, \emph{GRPO-Zero} trained on a self-generated pool frozen after the first round, a non-adaptive curriculum, and \emph{R-Zero}~\citep{rzero2025}. \textbf{Table~\ref{tab:main}} reports results across three backbone families. Our proposed \textbf{LURE} outperforms all trained baselines in all three environments. On PhantomWiki, it exceeds R-Zero by $15.0$ points and the frozen-pool control by an even larger margin. Since the frozen-pool control also improves over Base, self-generated data explains part of the gain, while the residual advantage over post-hoc band rejection reflects the contribution of adaptive task placement. On IFEval with the Qwen backbone, Base remains the strongest overall entry, and no self-play method surpasses the corrected zero-shot baseline. On ZebraLogic, \textbf{LURE} achieves smaller leads on both metrics. The larger separation reported previously resulted from gold-answer leakage rather than a genuine method difference.

\begin{figure*}[tbh]
  \centering
  \includegraphics[width=.85\textwidth]{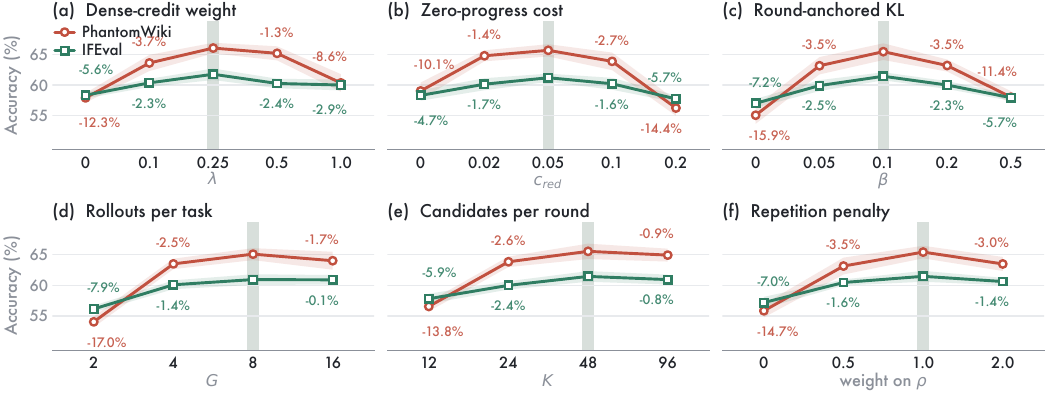}
  \caption{Hyperparameter sensitivity on Qwen2.5-7B unified model. Labels give each point's relative drop from its curve's best. }
  \label{fig:hparam}
\end{figure*}
\paragraph{Ablation Study.}
\textbf{Table~\ref{tab:ablation}} removes one component at a time, with each row's exact configuration specified in \textbf{App.~C}, and every removal lowers the mean, so each contributes to the full system. The magnitudes group by role: the three credit terms are the three smallest degradations, while every curriculum and stabilization ablation costs more than any of them, and the learned challenger is the largest at $-10.6$. The asymmetry follows from what each part controls, since the challenger decides which tasks carry a usable gradient at all, whereas the credit terms only reshape a gradient it has already supplied. Stabilization patterns with the curriculum rather than with the credit terms, which is what a mechanism protecting the loop from its own non-stationarity should do. Since the components interact through one shared verifier signal, these are contributions within the full system rather than independent effects.
\begin{table}[tbh]
  \centering
  \resizebox{\columnwidth}{!}{%
  \begin{tabular}{lccccr}
    \toprule
    Configuration & PhantomWiki & IFEval & ZebraLogic & Mean & $\Delta$\\
    \midrule
    \textbf{Full} & $\mathbf{65.8}$ & $\mathbf{61.2}$ & $16.9$ & $\mathbf{48.0}$ & ---\\
    \midrule
    \addlinespace[1pt]
    \multicolumn{6}{l}{\textit{Curriculum}}\\
    \quad \emph{w/o} learned challenger      & $40.0$ & $58.0$ & $14.1$ & $37.4$ & $\mathbf{-10.6}$\\
    \quad \emph{w/o} capture-frontier reward & $53.3$ & $58.0$ & $15.4$ & $42.2$ & $-5.8$\\
    \quad \emph{w/o} repetition penalty      & $55.8$ & $56.9$ & $14.6$ & $42.4$ & $-5.6$\\
    \addlinespace[3pt]
    \multicolumn{6}{l}{\textit{Pursuer credit}}\\
    \quad \emph{w/o} dense credit            & $57.5$ & $57.9$ & $15.1$ & $43.5$ & $-4.5$\\
    \quad \emph{w/o} zero-progress cost      & $59.2$ & $58.2$ & $15.8$ & $44.4$ & $-3.6$\\
    \quad \emph{w/o} planner credit          & $61.7$ & $59.5$ & $16.2$ & $45.8$ & $-2.2$\\
    \addlinespace[3pt]
    \multicolumn{6}{l}{\textit{Stabilization}}\\
    \quad \emph{w/o} round-anchored KL       & $55.0$ & $56.7$ & $13.4$ & $41.7$ & $-6.3$\\
    \quad \emph{w/o} stabilized recipe       & $52.5$ & $55.5$ & $12.9$ & $40.3$ & $-7.7$\\
    \bottomrule
  \end{tabular}}
  \caption{Ablation on the Qwen2.5-7B-Instruct unified model. $\Delta$ is the change in mean against the full system.}
  \label{tab:ablation}
\end{table}

\paragraph{Hyperparameter Sensitivity.}
\textbf{Fig.~\ref{fig:hparam}} evaluates the six method-specific hyperparameters: the dense-credit weight $\lambda$, zero-progress cost $c_{\mathrm{red}}$, round-anchored KL coefficient $\beta$, rollout group size $G$, candidate budget $K$, and the repetition-penalty scale, the coefficient on $\rho(x)$ that Eq.~\ref{eq:evader_reward} fixes to one. Performance is most sensitive to $\beta$ and $G$. Removing the round-anchored KL substantially reduces accuracy, while an excessively large $\beta$ also degrades performance, indicating a trade-off between stabilizing the co-evolving policies and preserving sufficient update flexibility. Reducing $G$ similarly causes a marked decline. Because the capture rate is estimated from $G$ rollouts, a small group yields a coarse estimate of the capture frontier and provides less reliable supervision for task selection. The remaining four parameters are comparatively stable around the selected settings, although extreme values can still reduce performance.

\paragraph{Evaluation Integrity.}
\textbf{Table~\ref{tab:audit}} examines whether the apparent baseline advantage on ZebraLogic reflects stronger reasoning or evaluation leakage. Panel~(a) disentangles access to the gold grid from the executor grammar. The reference condition reproduces the original result, while removing the gold grid reduces performance by $62$ points and changing the grammar moves it by at most $5.7$. The contrast locates the apparent advantage in prompt leakage rather than in the executor format or stronger reasoning. Panel~(b) applies the corrected evaluation protocol to PhantomWiki, where all trained methods improve and our method obtains the largest gain, further widening its margin. Because the leakage originates from a role-specific prompt rather than the benchmark answer key, reliable multi-role self-play evaluation requires auditing all role-conditioned prompts. \textbf{App.~B} lists the redacted fields.

\begin{table}[tbh]
  \centering
  \resizebox{0.85\columnwidth}{!}{
  \begin{tabular}{lccr}
    \toprule
    & Gold \emph{Kept} & Gold \emph{Removed} & $\Delta$\\
    \midrule
    \multicolumn{4}{l}{\textit{(a) ZebraLogic ($n{=}300$ Subset), R-Zero Round-5 Checkpoint}}\\
    \quad Old Grammar & $68.0$ & $6.0$ & $-62.0$\\
    \quad New Grammar & $62.3$ & $6.7$ & $-55.6$\\
    \addlinespace[4pt]
    \multicolumn{4}{l}{\textit{(b) PhantomWiki, Same Leak, Every Method}}\\
    \quad Base                 & $25.8$ & $25.0$ & $-0.8$\\
    \quad GRPO-Zero            & $35.8$ & $37.5$ & $+1.7$\\
    \quad R-Zero               & $49.2$ & $50.8$ & $+1.6$\\
    \quad \textbf{LURE (Ours)} & $60.8$ & $\mathbf{65.8}$ & $\mathbf{+5.0}$\\
    \bottomrule
  \end{tabular}
  }
  \caption{Evaluation integrity (\%). Both panels contrast the same two conditions on the same tasks.}
  \label{tab:audit}
\end{table}

\begin{figure}[tbh]
  \centering
  \includegraphics[width=.74\columnwidth]{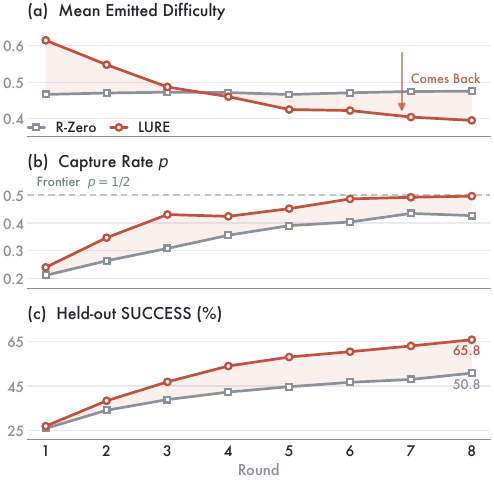}
  \caption{Anatomy of a chase on PhantomWiki per round.}
  \label{fig:chase}
\end{figure}

\paragraph{Curriculum Dynamics.}
\textbf{Fig.~\ref{fig:chase}} shows how \textbf{LURE} adapts task difficulty to the evolving pursuer on PhantomWiki. At round~1, the capture rate is well below the target frontier of $p=0.5$, indicating that the emitted tasks are initially too difficult. \textbf{LURE} subsequently lowers the mean difficulty, bringing the capture rate toward the frontier while held-out success increases from approximately $27\%$ to $65.8\%$. R-Zero instead maintains an approximately fixed emitted difficulty and achieves lower final capture and held-out performance. Thus, \textbf{LURE} maintains tasks near a fixed relative difficulty with respect to the current pursuer, rather than increasing difficulty according to a predefined schedule.

\paragraph{Curriculum Cost.}
\textbf{Table~\ref{tab:efficiency}} compares the rollout costs of post-hoc filtering and adaptive task placement. Both methods emit $144$ candidates per round. R-Zero spends $1152$ probe rollouts to retain $36$ tasks and discards the remaining $108$ before training, whereas \textbf{LURE} eliminates probing and trains on all $144$ tasks. Consequently, \textbf{LURE} trains on four times as many tasks at one-fifth the rollout cost per task, while reducing the total per-round cost from $1440$ to $1152$ rollouts.

\begin{table}[tbh]
  \centering
  \resizebox{.8\columnwidth}{!}{%
  \begin{tabular}{lcc}
    \toprule
    Rollout budget per round & R-Zero & \textbf{LURE (Ours)}\\
    \midrule
    \multicolumn{3}{l}{\textit{Selecting the curriculum}}\\
    \quad Candidates emitted         & $144$ & $144$\\
    \quad Probe rollouts             & $1152$ & $\mathbf{0}$\\
    \quad Candidates discarded       & $108$ & $\mathbf{0}$\\
    \addlinespace[3pt]
    \multicolumn{3}{l}{\textit{Training on it}}\\
    \quad Tasks entering training    & $36$ & $\mathbf{144}$\\
    \quad Training rollouts          & $288$ & $1152$\\
    \midrule
    Total rollouts                   & $1440$ & $\mathbf{1152}$\\
    Rollouts per training task       & $40.0$ & $\mathbf{8.0}$\\
    \bottomrule
  \end{tabular}}
  \caption{Curriculum cost. Per-round rollouts from candidate generation to training.}
  \label{tab:efficiency}
\end{table}

\begin{figure}[tbh]
  \centering
  \includegraphics[width=\columnwidth]{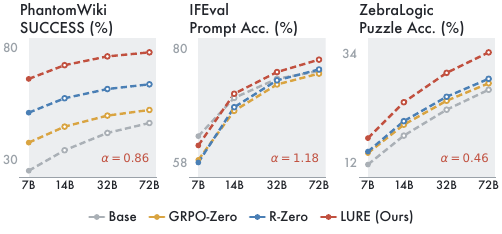}
  \caption{\textbf{Scaling behavior} (\%) across Qwen2.5-Instruct backbones from $7$B to $72$B. Filled markers are measured and hollow markers are projections of the power-law fit in \textbf{App.~C}, whose exponent $\alpha$ is shown for \textbf{LURE}.}
  \label{fig:scalinglaw}
\end{figure}

\paragraph{Scaling Behavior.}
\textbf{Fig.~\ref{fig:scalinglaw}} evaluates whether \textbf{LURE}'s advantage persists as the backbone scales from $7$B to $72$B based on Qwen2.5-Instruct series. The three environments exhibit distinct scaling patterns. On PhantomWiki, all methods show diminishing gains at larger scales, while \textbf{LURE} maintains a substantial margin over the trained baselines. On IFEval, the methods converge toward a similar performance range: Base performs best at $7$B, whereas \textbf{LURE} surpasses it from $14$B onward. On ZebraLogic, \textbf{LURE}'s margin increases

\paragraph{OOD Generalization.}
We further evaluate OOD generalization of the unified model on nine benchmarks across competition mathematics (\textbf{MATH500}~\citep{hendrycks2021math,lightman2024lets}, \textbf{AIME 2024}, \textbf{AIME 2025}~\citep{aime}, and \textbf{OlympiadBench}~\citep{olympiadbench}), rule-based and general reasoning (\textbf{KOR-Bench}~\citep{korbench}, \textbf{LiveBench-Reasoning}, and \textbf{LiveBench-IF}~\citep{livebench}), and relational or spatial inference (\textbf{CLUTRR}~\citep{clutrr} and \textbf{StepGame}~\citep{stepgame}). \textbf{Table~\ref{tab:transfer}} reports results under a raw single-shot prompt without the planner or executor role, isolating the capabilities retained by the trained backbone, with the per-benchmark protocol in \textbf{App.~B}. No method shows consistent gains over Base across individual benchmarks, indicating that the in-domain improvements do not transfer uniformly. At the aggregate level, however, \textbf{LURE} is the only trained method to improve over Base, increasing the overall mean by $0.6$ points, whereas GRPO and GRPO-Zero decrease by $0.1$ and $1.0$ points and R-Zero matches Base. Our \textbf{LURE} also achieves the highest aggregate score in all three benchmark families, indicating stronger OOD retention despite modest and benchmark-dependent gains.

\begin{table}[tbh]
  \centering
  \resizebox{.93\columnwidth}{!}{%
  \begin{tabular}{lcccccc}
    \toprule
    Benchmark & $n$ & Base & GRPO & GRPO-Zero & R-Zero & \textbf{LURE}\\
    \midrule
    MATH500             & $100$ & $77.0$ & $78.0$ & $77.0$ & \tb{81.0} & \ts{79.0}\\
    AIME24              & $30$  & \ts{13.3} & \ts{13.3} & $10.0$ & \ts{13.3} & \tb{16.7}\\
    AIME25              & $30$  & \tb{10.0} & \tb{10.0} & \tb{10.0} & \tb{10.0} & \tb{10.0}\\
    OlympiadBench       & $300$ & \tb{43.0} & \ts{42.3} & $41.7$ & $41.0$ & $41.7$\\
    KOR-Bench           & $250$ & \ts{37.6} & $37.2$ & $36.8$ & $36.0$ & \tb{39.6}\\
    LiveBench-Reasoning & $100$ & \tb{26.3} & \ts{26.0} & $25.0$ & $25.8$ & $25.0$\\
    LiveBench-IF        & $200$ & \ts{46.1} & $46.0$ & $45.5$ & \tb{46.5} & \ts{46.1}\\
    CLUTRR              & $300$ & $32.7$ & $32.7$ & $32.0$ & \ts{33.0} & \tb{33.7}\\
    StepGame            & $200$ & \tb{20.5} & \tb{20.5} & \ts{20.0} & \tb{20.5} & \tb{20.5}\\
    \midrule
    \textit{Mathematics}& $460$ & $35.8$ & $35.9$ & $34.7$ & \ts{36.3} & \tb{36.9}\\
    \textit{Rule-based} & $550$ & \ts{36.7} & $36.4$ & $35.8$ & $36.1$ & \tb{36.9}\\
    \textit{Relational} & $500$ & $26.6$ & $26.6$ & $26.0$ & \ts{26.8} & \tb{27.1}\\
    \midrule
    Mean                 &       & $34.1$ & $34.0$ & $33.1$ & \ts{34.1} & \tb{34.7}\\
    $\Delta$ vs Base     &       & ---    & $-0.1$ & $-1.0$ & \ts{0.0} & \tb{+0.6}\\
    \bottomrule
  \end{tabular}}
  \caption{OOD generalization (\%) with a raw single-shot prompt and no planner or executor role, on the Qwen2.5-7B-Instruct unified model. \colorbox{bestbg}{\textbf{Best}} and \colorbox{secbg}{\underline{second best}} per row.}
  \label{tab:transfer}
\end{table}

\section{Conclusion}
We propose \textbf{LURE}, which recasts zero-data self-play as a pursuit-evasion game between a challenger that places tasks and a planner-executor solver that hunts them down. The challenger is trained on a capture-frontier reward that peaks at a capture rate of one half, while the solver receives capture-anchored dense process credit under a round-anchored KL that holds both stable. Across three verifiable reasoning environments and backbone families, \textbf{LURE} outperforms baselines under unified and specialist settings and attains stronger aggregate OOD zero-shot accuracy across nine benchmarks, though the gains remain modest and benchmark-dependent.

\bibliography{references}

\end{document}